\documentclass[letterpaper 10pt, conference]{IEEEtran} 

\IEEEoverridecommandlockouts                              

\usepackage[numbers, sort&compress]{natbib}
\usepackage[utf8]{inputenc} 
\usepackage[T1]{fontenc}    
\usepackage{hyperref}       
\usepackage{url}            
\usepackage{booktabs}       
\usepackage{amsfonts}       
\usepackage{nicefrac}       
\usepackage{microtype}      
\usepackage{graphicx}
\usepackage{amsmath,amssymb,amsfonts}
\usepackage[acronym]{glossaries}
\usepackage{comment}
\usepackage{xcolor}
\usepackage{enumerate}
\usepackage[capitalise]{cleveref}
\usepackage{caption}
\usepackage{svg}
\usepackage{listings}
\usepackage{siunitx}
\usepackage[section]{placeins}
\usepackage[english]{babel}
\usepackage{booktabs}
\usepackage{multirow}
\usepackage{cancel}
\usepackage[normalem]{ulem}
\useunder{\uline}{\ul}{}
\usepackage{eso-pic}
\usepackage{tikz}
\usepackage{amsmath}
\usepackage{cuted}

\newcommand{\rev}[1]{#1}
\usepackage[normalem]{ulem} 
\newcommand{\revdel}[1]{} 

\title{\LARGE \bf \vspace{6mm}
Learning-Based On-Track System Identification for \\Scaled Autonomous Racing in Under a Minute
}

\author{Onur Dikici\IEEEauthorrefmark{1}, Edoardo Ghignone\IEEEauthorrefmark{2}, Cheng Hu\IEEEauthorrefmark{3}, Nicolas Baumann\IEEEauthorrefmark{2},\\ Lei Xie\IEEEauthorrefmark{3}, Andrea Carron\IEEEauthorrefmark{2}, Michele Magno\IEEEauthorrefmark{2}, and Matteo Corno\IEEEauthorrefmark{1} \\
\thanks{\IEEEauthorrefmark{1}Onur Dikici and Matteo Corno are associated with the Dipartimento di Elettronica, Informazione e Bioingegneria, Politecnico di Milano}
\thanks{\IEEEauthorrefmark{2}Edoardo Ghignone, Nicolas Baumann, Andrea Carron, and Michele Magno are associated with ETH Zürich.}%
\thanks{\IEEEauthorrefmark{3}Cheng Hu and Lei Xie are associated with the Department of Control Science and Engineering, Zhejiang University.
}
\thanks{\emph{Onur Dikici, Edoardo Ghignone contributed equally to this work. (Corresponding author: Onur Dikici.)}
}
}
\begin{document}
\newacronym{mpc}{MPC}{Model Predictive Control}
\newacronym{mpcc}{MPCC}{Model Predictive Contouring Controller}
\newacronym{rl}{RL}{Reinforcement Learning}
\newacronym{mlp}{MLP}{Multilayer Perceptron}
\newacronym{forl}{FoRL}{Foundations of Reinforcement Learning}
\newacronym{ml}{ML}{Machine Learning}
\newacronym{sb3}{SB3}{Stable Baselines 3}
\newacronym{sac}{SAC}{Soft Actor Critic}
\newacronym{ppo}{PPO}{Proximal Policy Optimization}
\newacronym{ai}{AI}{Artificial Intelligence}
\newacronym{nn}{NN}{Neural Network}
\newacronym{sota}{SotA}{State-of-the-Art}
\newacronym{esc}{ESC}{Electronic Speed Controller}
\newacronym{ros}{ROS}{Robot Operating System}
\newacronym{imu}{IMU}{Inertial Measurement Unit}
\newacronym{ekf}{EKF}{Extended Kalman Filter}
\newacronym{slam}{SLAM}{Simultaneous Localization And Mapping}
\newacronym{sdc}{SDC}{Self Driving Cars}
\newacronym{obc}{OBC}{On Board Computer}
\newacronym{qp}{QP}{Quadratic Programming}
\newacronym{uav}{UAV}{Unmanned Aerial Vehicles}
\newacronym{cg}{CG}{Center of Gravity}
\newacronym{em}{EM}{Expectation Maximization}
\newacronym{rms}{RMS}{Root Mean Square}
\newacronym{rmse}{RMSE}{Root Mean Square Error}
\newacronym{mse}{MSE}{Mean Square Error}
\newacronym{nrmse}{NRMSE}{Normalized Root Mean Square Error}
\newacronym{map}{MAP}{Model- and Acceleration-based Pursuit}
\newacronym{pd}{PD}{Proportional-Derivative}
\newacronym{lut}{LUT}{Lookup Table}
\newacronym{nls}{NLS}{Nonlinear Least Squares}
\newacronym{gp}{GP}{Gaussian Process}
\newacronym{elm}{ELM}{Extreme Learning Machine}
\newacronym{ads}{ADS}{Autonomous Driving Systems}
\newacronym{uslam}{USLAM}{Ultimate Simultaneous Localization and Mapping}

\pdfcompresslevel=9

\maketitle

\begin{strip}
\vspace{-3.55cm}
\centering
\includegraphics[width=\textwidth]{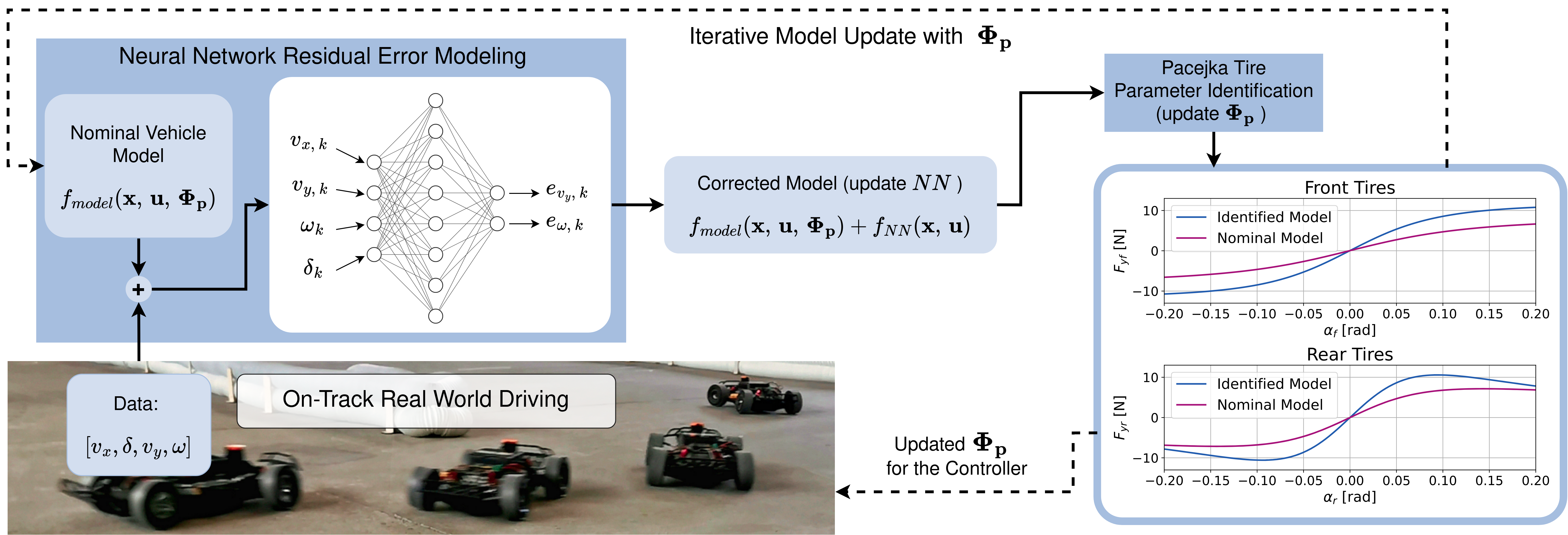}
\captionof{figure}{Schematic representation of the learning-based on-track system identification method for autonomous racing. The nominal vehicle model is corrected using a residual \gls{nn} error model\rev{, trained} \revdel{based}on data collected during on-track\rev{,} real-world driving. \revdel{The corrected model}\rev{Pacejka parameters} are iteratively updated\revdel{, enhancing} \rev{to enhance} the \rev{overall} accuracy of the \revdel{tire parameter identification through the Pacejka model}\rev{vehicle model}.}
\label{fig:graphical_abstract}
\vspace{-0.275cm}
\end{strip}

\thispagestyle{empty}
\pagestyle{empty}

\begin{abstract}

Accurate tire modeling is crucial for optimizing autonomous racing vehicles\rev{, as \gls{sota} model-based techniques rely on precise knowledge of the vehicle's parameters}, yet system identification in dynamic racing conditions is challenging due to varying track and tire conditions. Traditional methods require extensive operational ranges, often impractical in racing scenarios. 
\gls{ml}-based methods, while improving performance, struggle with generalization and depend on accurate initialization. This paper introduces a novel on-track system identification algorithm, incorporating a \gls{nn} \rev{for error correction, which is then employed} \revdel{with a Pacejka tire model. The proposed approach trains a \gls{nn} for error correction and then employs it} for traditional system identification with virtually generated data. 
Crucially, the process is iteratively reapplied, with tire parameters updated at each cycle, leading to notable improvements in accuracy in tests on a scaled vehicle.
Experiments show that it is possible to learn a \rev{tire} model without prior knowledge with only 30 seconds of driving data, and 3 seconds of training time\revdel{, corresponding to six iterations. Furthermore, as only 0.5 seconds are needed per iteration, the model can also be refined in real-time}.
This method demonstrates \revdel{significantly} greater \rev{one-step prediction} accuracy than \rev{the baseline \gls{nls} method} under noisy conditions, achieving \rev{a 3.3}x lower \gls{rmse}, and yields \rev{tire} models with comparable accuracy to traditional steady-state system identification\revdel{, while also significantly simplifying the process by enabling }\rev{. Furthermore, unlike steady-state methods requiring large spaces and specific experimental setups, the proposed approach identifies tire parameters directly on a race track}\revdel{real-time, on-track system identification} in dynamic racing environments.

\begin{IEEEkeywords}
    Field Robots, Wheeled Robots, Machine Learning for Robot Control
\end{IEEEkeywords}
\end{abstract}

\section{INTRODUCTION}

Autonomous racing offers a dynamic, high-speed environment that requires high performance under extreme conditions, making it ideal for testing the limits of \gls{ads} in scenarios similar to potential edge cases in real-world driving \cite{forzaeth}. It provides a controlled yet demanding setting for advancing autonomous driving technologies without risking public traffic safety \cite{ar_survey}. The challenging nature of racing drives the development of advanced data-driven algorithms that perform reliably at high speeds and handle complex dynamics, such as operating at the edge of friction. Additionally, racing supplies measurable metrics like lap time and tracking error to evaluate the performance of an \gls{ads}, addressing the difficulties of quantifying performance in general autonomous driving \cite{ar_survey}.

A primary challenge in autonomous racing is to complete a racetrack at high speeds \cite{ar_survey, forzaeth}, which poses a demanding task for autonomous control systems due to the highly non-linear behavior of the car, when tire dynamics exceed the linear operating range and the vehicle must operate at the edge of traction \cite{pacejka1992magic, map, xue2024learningMPC, chrosniak2024deepdynamics}. 
Thus, accurately tracking trajectories at cornering speeds that saturate tire friction capacity is a central and highly challenging control task in autonomous racing \cite{amz_fullstack, indyautonomous, ar_survey}.

To achieve said level of high performance, model-based controllers must utilize knowledge of \rev{vehicle} model dynamics to compute effective control signals \cite{liniger_mpcc, map}. These types of controllers require prior knowledge of tire dynamics, which are notoriously difficult to estimate in field conditions \cite{map}. Thus, accurate system identification of the entire model, with a particular focus on the complex tire dynamics, is essential for enabling high-performance model-based controllers \cite{chrosniak2024deepdynamics}.

Typically, this type of system identification is performed prior to a race through meticulous tire characterization experiments. These experiments are time-consuming and require a large, open space with a surface similar to the actual racetrack \cite{ss_sys_id, map, raji2022motion}. \revdel{Consequently, enabling model-based controllers in varying environmental contexts, as encountered in real-world racing events, demands accurate tire parameter identification and operational robustness. Unanticipated variables, such as changes in surface traction at an unknown racetrack, must be considered. Therefore, it is crucial that system identification processes allow for operational simplicity \cite{forzaeth} and can be deployed without the need for large open spaces, which might not be available before a race.}\rev{Enabling model-based controllers in real-world racing requires accurate tire parameter identification and robustness to variables like surface traction changes. System identification must be simple and deployable without large open spaces, which are often unavailable before a race \cite{forzaeth}.}
Autonomous driving literature has identified two primary types of system identification methods:
\begin{enumerate}[I]
    \item \textbf{Off-Track Identification:} These methods require steady-state conditions. They are accurate because system identification is conducted through steady-state experiments, providing a simplified dynamics environment with ideal conditions \cite{ss_sys_id, map, raji2022motion, seong2023model}. Additionally, because the vehicle is in a steady-state, there is no need to deal with differentiating noisy measurements. However, they necessitate large, open spaces similar to the actual racetrack surface, which may not always be available, especially given the variable conditions within the nature of racing.
    \item \textbf{On-Track Identification:} These methods do not require large spaces for steady-state conditions but are more challenging due to the vehicle not being in a steady-state. Optimization methods, such as \gls{nls} \cite{vehicle_dynamics_book, brunner2017repetitive}, are used to perform this type of identification but can be brittle and fail to produce reasonable results in practice. Different optimization methods can then successfully produce estimates, such as the non-convex program presented in \cite{bodmer2024optimization} or the method in \cite{learning_mpc}, which needs a good initial estimate of the tire parameters. 
\end{enumerate}

To address the limitations of previous system identification methods, this paper proposes a novel learning-based on-track system identification method. This method not only improves the accuracy of parameter identification but also significantly enhances the operational simplicity of tire parameter characterizations. It eliminates the need for space- and time-consuming identification maneuvers and overcomes the brittleness of conventional \gls{nls}-based on-track identification methods by leveraging the efficiency of data-driven \gls{nn} learning.
 The contributions of this work are summarized as follows:
\begin{enumerate}[I]
    \item \textbf{On-Track Identification:} The proposed method identifies tire characteristics directly on the track, eliminating the need for steady-state identification experiments in large open spaces \cite{ss_sys_id}. This approach allows the identification process to be conducted on the actual track being used, removing the logistical challenges of finding a similar surface prior to the race and providing high operational simplicity. Furthermore, thanks to the combined method, the typical overfitting problem of \gls{ml} techniques is avoided, as tire parameters are eventually obtained from our technique.
    \item \textbf{Robustness:} Our combined method using a \gls{nn}-based learning scheme with traditional system identification demonstrates high robustness to noise encountered in real-life deployments. Compared to \gls{nls}-based on-track identification methods, our approach achieves up to \rev{3.3} times greater accuracy under noisy conditions in simulation and handles the noise experienced in real-world robotic systems, whereas conventional methods often fail to identify parameters outside of simulation environments.
    \item \textbf{Open-Source:} The proposed identification method is fully integrated into an open-source full-stack implementation \cite{forzaeth}, enhancing reproducibility and extensibility. \rev{Code available at: \href{https://github.com/ForzaETH/On-Track-SysID}{\url{https://github.com/ForzaETH/On-Track-SysID}}.}
\end{enumerate}

\section{RELATED WORK}

In system identification for autonomous vehicles, steady-state techniques like \cite{ss_sys_id, raji2022motion} face limitations due to their requirement for large free areas for circular motion. 
The work presented in \cite{raji2022motion} uses high-fidelity simulators and high-quality initial estimates, while other approaches rely on provided ground truth tire data to learn \cite{seong2023model}. 
On-track methods \cite{brunner2017repetitive, vicente2021linear, learning_mpc, bodmer2024optimization, chrosniak2024deepdynamics}, which utilize data collected during driving along the same track destined for a competition, present a more practical alternative.
Among these methods, classical approaches like \gls{nls} are some of the suggested ones \cite{vehicle_dynamics_book}, and they are implemented online for instance in \cite{brunner2017repetitive}.
However, these methods could be sensitive to noisy data, and \cite{vicente2021linear} shows that simple data-driven techniques can outperform least-squares-based system identification on real data.

Different approaches employed \glspl{gp} for efficiently exploiting real data, such as \cite{bayesrace}, where a kinematic model is extended with a \gls{gp} and iteratively updates to learn model mismatch in simulation, or \cite{NAGY2023ensemble} where the authors present a method to utilize an ensemble of \glspl{gp} to be able to model varying friction.
Both these models however suffer from computational limitations, especially with the need to continuously update the models, and, to mitigate computational costs, different works utilize sparse \gls{gp} regression, such as \cite{cautious_nmpc_gp, learning_mpc}. 
Despite these improvements, their methods remain untested under significant model mismatches, as they often start with an initially accurate physical model.

Furthermore, \gls{nn} techniques have been proposed to address the computational cost, such as in \cite{extreme_learning_machine} where an  \gls{elm} is used to successfully identify dynamic vehicle models in simulation, or in \cite{end-to-end} where an end-to-end \gls{nn} is proposed for learning vehicle dynamics. This last work, however, highlights how these \gls{ml} techniques can fail to generalize across diverse conditions and would require extensive data gathering to be deployed in online applications.
\rev{Recent work explored the adaptability with \gls{ml} models more extensively, such as with Continual-MAML \cite{tsuchiya2024onlineadaptationlearnedvehicle}, or with different architectures such as LSTMs \cite{kalaria2024agilemobilityrapidonline} or Transformers \cite{xiao2024anycaranywherelearninguniversal}.}

This work builds on top of this last direction, aiming to exploit the efficiency in data processing of \glspl{nn} and to address two main issues with the hybrid architecture joining \gls{ml} and traditional system identification.
Firstly, we show how our method can be more practical than end-to-end learning techniques \cite{extreme_learning_machine, end-to-end} and steady-state system identification procedures \cite{ss_sys_id, raji2022motion} by learning with less than a minute of training data and not requiring the design of specific experiments or significantly large spaces. 
Furthermore, we address the need for adaptable models, already highlighted by \cite{brunner2017repetitive, NAGY2023ensemble}, by presenting a lightweight model that can adapt to new data in \revdel{only 10 seconds}\rev{at most 3 seconds of training time}, effectively enabling online adaptation on a fully onboard setup.
\rev{Crucially, our method differs from adaptive \gls{ml} techniques for learning dynamic models \cite{tsuchiya2024onlineadaptationlearnedvehicle, kalaria2024agilemobilityrapidonline, xiao2024anycaranywherelearninguniversal} in a key way: rather than only updating the \gls{nn}, our approach iteratively updates both the tire model and the \gls{nn}. Jointly updating both models allows for accurate dynamic representation even with a relatively simple model comprising less than 60 parameters. Furthermore, it also paves the way for future work intersecting novel techniques \cite{tsuchiya2024onlineadaptationlearnedvehicle, kalaria2024agilemobilityrapidonline, xiao2024anycaranywherelearninguniversal} and traditional empirical methods such as the Pacejka tire model.}

\section{SYSTEM IDENTIFICATION}

In this section, we provide an overview of our methodology, which proposes a \gls{nn} model to capture the model mismatch. Although modeling the mismatch with a data-driven approach may fail to represent out-of-distribution data, simulating and generating predictions with in-distribution data enables the extraction of the underlying Pacejka parameters embedded within the combined \gls{nn} and vehicle model. Furthermore, this process is done iteratively to gradually improve the \rev{accuracy of the identified tire} model\revdel{'s accuracy} with each iteration.

\subsection{Vehicle Model} \label{subsec:vehicle_model}

\begin{figure}[h!]
    \centering
    \includegraphics[width=\columnwidth]{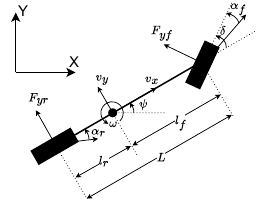}
    \caption{Dynamic single-track model showing lateral forces (\(F_{yf}\), \(F_{yr}\)), slip angles (\(\alpha_f\), \(\alpha_r\)), velocities (\(v_x\), \(v_y\)), steering angle (\(\delta\)), yaw rate (\(\omega\)), angle with respect to the  positive X axis ($\psi$), and X-Y coordinate system. The symbols $l_r\rev{,}\,l_f$ indicate, respectively, the distance from the center of gravity to the rear and front axle, and $L$ indicates the full wheelbase length. \rev{\textit{Note: This image is illustrative; slip angles are exaggerated for visualization purposes.}}}   
    \label{fig:dynamic_bicycle_model}
\end{figure}

The vehicle model used in this work \rev{illustrated in \Cref{fig:dynamic_bicycle_model}}, is the dynamic single-track model, a commonly used vehicle model in the field of autonomous racing as it effectively models tire-road interactions \rev{in steady-state conditions }\cite{modeling_survey}. This model \rev{simplifies the vehicle to one wheel per axis and decouples longitudinal and lateral dynamics. While this approach captures the essential vehicle behavior, it relies on a steady-state tire model that may not fully account for the dynamic response of tires during rapid maneuvers.}\revdel{ is illustrated in Fig. It assumes the vehicle as a simplified representation where only one wheel per axle is considered and the longitudinal and lateral dynamics are decoupled.} The vehicle is considered to be a rigid body with a mass \(m\) and an inertia \(I_z\) around the z-axis of the center of gravity. \(l_f\) and \(l_r\) represent the distance of the front and rear axle from the center of gravity, respectively.

The equation of motion for the lateral model can be expressed as follows:
\begin{equation} \label{eq:vehicle_model}
\begin{aligned}
\dot{v}_y & =\frac{1}{m}\left(F_{yr}+F_{yf} \cos \delta-m v_x \omega\right) \\
\dot{\omega} & =\frac{1}{I_z}\left(F_{yf} l_f \cos \delta-F_{yr} l_r\right).
\end{aligned}
\end{equation}
To accurately model the lateral forces \(F_{yf}\) and \(F_{yr}\), the Pacejka \textit{Magic Formula} \cite{pacejka1992magic} is used in its general form
\begin{equation}
F_{yi} = D_i \sin(C_i \arctan(B_i \alpha_i - E_i (B_i \alpha_i - \arctan(B_i \alpha_i)))).
\end{equation}
where \(i\) is an element of \( \{f, r\} \), representing the front and rear tires respectively. \rev{Due to the low center of gravity of the scaled car used in this work, longitudinal and lateral load transfers are neglected.} The slip angles are also computed as follows:
\begin{equation} \label{eq:slip_angles}
\begin{aligned}
\alpha_f = \delta - \arctan\left(\frac{v_y + l_f \omega}{v_x}\right) \\
\alpha_r = - \arctan\left(\frac{v_y - l_r \omega}{v_x}\right)
\end{aligned}.
\end{equation}
This work focuses solely on lateral vehicle dynamics, specifically the identification of Pacejka model parameters \(\mathbf{\Phi_p} = [B_f, C_f, D_f, E_f, B_r, C_r, D_r, E_r]\). Accordingly, we treat \(\mathbf{x} = [v_y,\omega]\) as states, and \(\mathbf{u} = [v_x,\delta]\) as inputs. Using Euler integration, the discrete-time vehicle model becomes:
\begin{equation} \label{eq:simplified_model_discrete}
\begin{aligned}
v_{y,k+1} &= v_{y,k} + \frac{1}{m}\left(F_{yr} + F_{yf} \cos \delta - m v_x \omega\right) T_s \\
\omega_{k+1} &= \omega_{k} + \frac{1}{I_z}\left(F_{yf} l_f \cos \delta - F_{yr} l_r\right) T_s,
\end{aligned}
\end{equation}
where \(T_s\) is the sampling time.

\subsection{Model Mismatch Problem} \label{subsec:model_error}

There are instances when the identified Pacejka parameters \(\mathbf{\Phi_p}\) may no longer be accurate. This can result from changes in the racing surface---as Pacejka parameters are surface-dependent---as well as from tire wear and variations in tire temperature. Consequently, there may be a discrepancy between the predicted subsequent states \(\mathbf{\hat{x}}_{k+1}\) and the measured subsequent states \(\mathbf{x}_{k+1}\). This discrepancy is known as model mismatch and is expressed as follows:

\begin{equation} \label{eq:model_mismatch}
\mathbf{e}_k = \mathbf{x}_{k+1} - \mathbf{\hat{x}}_{k+1},
\end{equation}
where \(\mathbf{\hat{x}}_{k+1} = [\hat{v}_{y,k+1}, \hat{\omega}_{k+1}]\) is calculated using \eqref{eq:simplified_model_discrete} with the nominal Pacejka parameters \(\mathbf{\Phi_p}\).


\subsection{Data Collection \& Data Processing}\label{subsec:data_collection_preperation}

\subsubsection{Data Collection}

Data is collected by driving the vehicle on a track with a Pure Pursuit controller \cite{pure_pursuit}, using the implementation from \cite{forzaeth}.
Crucially, such a controller does not require any \rev{tire or vehicle} model knowledge apart from the wheelbase length $L$, making it apt as a starting baseline.
A single collected data tuple for \gls{nn} training is identified with $\mathbf{\mathcal{D}}^{train} = [ v_{x}, v_{y}, \omega, \delta]$.

\subsubsection{Data Processing}
\revdel{After collecting the data, several processing steps are necessary to prepare it for \gls{nn} training. 
Firstly, the vehicle state estimation data may contain noise due to various factors such as measurement errors, road irregularities, and other environmental factors. To sanitize this data, we used a noncausal low-pass filter. This filter is crucial because it avoids phase delays and smooths the data, making it more reliable for \gls{nn} training. 
Secondly, we assumed the vehicle dynamics to be perfectly symmetric and thus extended the dataset with mirrored lateral velocity, yaw rate, and steering angle while keeping the longitudinal velocity unchanged.
This technique not only increases the amount of training data but also removes the intrinsic bias in one type of cornering direction present in every racing track due to their looping nature.}
\rev{After collecting the data, preprocessing is performed to prepare it for \gls{nn} training. First, a noncausal low-pass filter is applied to handle noise from measurement errors and environmental factors, ensuring smooth and reliable data without introducing phase delays. Second, assuming symmetric vehicle dynamics, the dataset is augmented by mirroring lateral velocity, yaw rate, and steering angle, while keeping longitudinal velocity unchanged. This augmentation increases the data volume and eliminates the cornering bias inherent to racing tracks.}

\subsection{\acrlong{nn} for Identification} \label{subsec:nn}


The \gls{nn} is tasked with learning and correcting the model error to improve the nominal vehicle \revdel{dynamics}model. The targets of the \gls{nn} are the prediction errors \(\mathbf{e}_k = [ e_{v_{y,k}},\ e_{\omega_k}]\) for the subsequent step's lateral velocity and yaw rate. 
These errors are determined\revdel{as from} \rev{as described in} \eqref{eq:model_mismatch}.

The inputs for the \gls{nn}\revdel{is} \rev{are} chosen to be \([\mathbf{x}_k, \mathbf{u}_k] =[v_{x,k}, v_{y,k}, \omega_k, \delta_k]\) since the model error depends on these variables in the determined vehicle model. In summary, the \gls{nn} models the prediction error of a nominal vehicle model based on these inputs. This function learned by the \revdel{neural network}\rev{\gls{nn}} is denoted as follows:
\begin{equation} \label{eq:nn_function}
\begin{aligned}
\mathbf{e}_k = f_{NN}(v_{x,k}, v_{y,k}, \omega_k, \delta_k),
\end{aligned}
\end{equation}
representing the prediction of the error \(\mathbf{e}_k\) given the inputs.

\subsection{Parameter Identification with the Corrected \rev{Vehicle}  Model} \label{subsec:param_id_combined_model}
The \glspl{nn} performance is constrained by the limited training data available, making globally accurate behavior challenging and struggling with sparse, non-existent, or significantly different inputs from the training data (out-of-distribution). To address those limitations, we identify Pacejka parameters, as the Pacejka \rev{tire} model accurately represents vehicle dynamics across diverse conditions. By simulating the corrected \rev{vehicle} model with a range of well-represented input variables, we ensure that the \gls{nn} operates within in-distribution data, leading to reliable predictions.
\subsubsection{\rev{The} Corrected \rev{Vehicle} Model}\label{subsec:combined_model}
\begin{figure}[h!]
    \centering
    \includegraphics[width=\columnwidth]{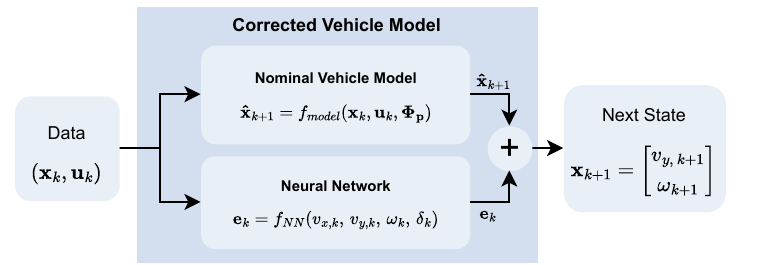}
    \caption{\rev{The} corrected \rev{vehicle }model structure combining a nominal vehicle model and a \gls{nn} to predict the next state variables \(v_{y,k+1}\) and \(\omega_{k+1}\). Where \(f_{\text{model}}(\mathbf{x}_k, \mathbf{u}_k, \mathbf{\Phi_p})\) represents the state transition function based on \eqref{eq:simplified_model_discrete}, with  \(\mathbf{x}_k\) and \(\mathbf{u}_k\) being the state and input vectors, respectively, and \(\mathbf{\Phi_p}\) are the Pacejka parameters.} 
    \label{fig:combined_model}
\end{figure}

The corrected \rev{vehicle} model leverages both the \gls{nn} and the nominal vehicle model, similarly to \cite{xue2024learningMPC}. The nominal model provides a baseline prediction, while the \gls{nn} corrects residual errors. This structure is shown in \Cref{fig:combined_model}.

The corrected next states \([v_{y,k+1}, \omega_{k+1}]\) are calculated as follows:
\begin{equation} \label{eq:corr_state_calc}
\begin{aligned}
\mathbf{x}_{k+1} = \mathbf{\hat{x}}_{k+1} + \mathbf{e}_k,
\end{aligned}
\end{equation}
where \(\mathbf{\hat{x}}_{k+1}\) is the predicted next state from the nominal vehicle model using \eqref{eq:simplified_model_discrete}, and \(\mathbf{e}_k\) is the residual error predicted by the \gls{nn} using \eqref{eq:nn_function}.

\subsubsection{Virtual Steady-State Data Generation\revdel{with the Corrected Model}}\label{subsec:data_gen}

To identify Pacejka parameters, we first generate virtual steady-state data by simulating the corrected \rev{vehicle} model at a constant longitudinal speed. The chosen speed is the average longitudinal velocity of the training data, ensuring the \gls{nn} performs confidently due to its familiarity with this type of data. During the simulation, the steering angle is gradually increased from 0 to 0.4 radians linearly, based on the operational limits of the target platform, a 1:10 scaled vehicle. By incrementing the steering angle in this controlled manner, we can observe and record the vehicle's response under steady-state conditions.

The simulation initializes the states $\mathbf{x}_0 = [v_{y,0}\ \omega_{0}] = [0\ 0]$. The prior next state predictions, $\mathbf{\hat{x}}_{k+1}$, from the nominal model are corrected with the \gls{nn}'s error prediction, $\mathbf{e}_{k}$, using the current states and inputs as described in \eqref{eq:corr_state_calc}. This process generates the entire steady-state data over 10 seconds with a time step of 0.02 seconds.

\subsubsection{Pacejka Model Parameters Identification} \label{subsec:model_id}

Since the generated data is in steady-state, we assume $\dot{v}_y = 0$ and $\dot{\omega} = 0$. This allows us to accurately compute the lateral forces $F_{yr}$ and $F_{yf}$ using the states and inputs:

\begin{equation} \label{eq:simplified_forces}
\begin{aligned}
F_{yr} = \frac{ml_f}{l_f + l_r} v_x \omega \\
F_{yf} = \frac{ml_r}{l_f + l_r} \frac{v_x \omega}{\cos(\delta)}
\end{aligned}.
\end{equation}

The slip angles $\alpha_f$ and $\alpha_r$ are calculated using \eqref{eq:slip_angles}. We then fit the Pacejka model to the force vs. slip angle data $(F_{yf}, \alpha_f)$ and $(F_{yr}, \alpha_r)$ through least squares regression \cite{ss_sys_id, map}.

\subsection{Iterative Learning} \label{subsec:iterative_learning}

\begin{figure}[ht]
    \centering
\includegraphics[width=\columnwidth]{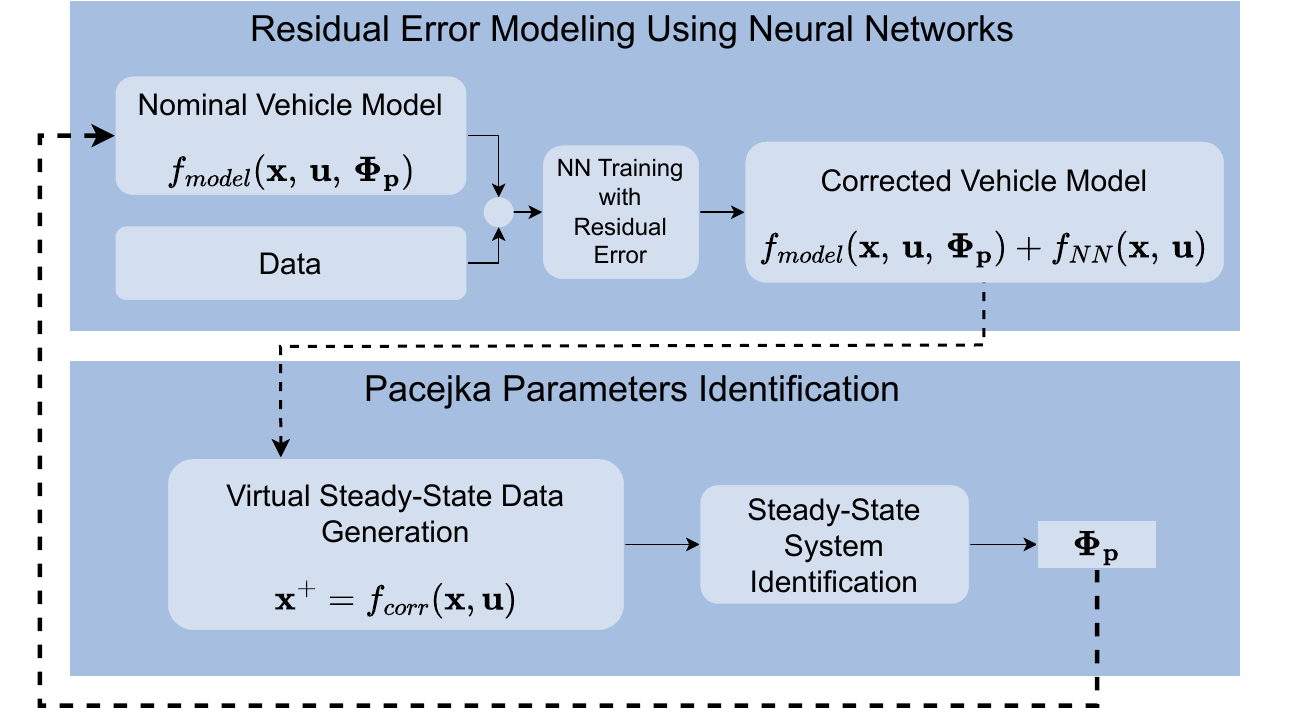}
    \caption{Scheme of the iterative approach: Starting with residual error modeling and corrected \rev{vehicle} model formation, followed by steady-state system identification using virtual data from the corrected \rev{vehicle} model. Identified parameters are then used as a new nominal model for learning residual errors, repeating iteratively until convergence.}
    \label{fig:IterativeApproach}
\end{figure}

Even though the identified Pacejka parameters improve the model, a mismatch may still persist if the initial mismatch is high. To address this, we iteratively refine the model by using the identified parameters as a new nominal model, learning its residuals, and repeating the process until the model converges. \rev{The convergence is determined empirically by observing that further iterations do not yield significant improvements in the identified Pacejka models.} This approach ensures that the \gls{nn} \revdel{refines its predictions}\rev{continuously learns residual errors} based on progressively more accurate Pacejka parameters,\revdel{continuously} improving the \rev{overall vehicle} model accuracy. The scheme for this iterative approach is demonstrated in \Cref{fig:IterativeApproach}. 

One significant advantage of this iterative approach is that it removes the dependency on having an initially accurate nominal model. Even if the initial model has a high mismatch, the iterative process allows for continuous improvement of the nominal model's accuracy until it converges. By identifying Pacejka parameters and \rev{re-initializing}\revdel{resetting} the \gls{nn}, we prevent overfitting and progressively refine the model's accuracy.

\section{EXPERIMENTAL RESULTS}
In this section, we present the metrics utilized for assessing \revdel{system's}\rev{the} accuracy and performance \rev{of the identified models} and then show the results in simulation and on the physical 1:10 scaled autonomous racing platform.

\subsection{Model Architecture}
The \gls{nn} architecture used is a feed-forward \gls{mlp} with an input layer of 4 neurons, one hidden layer of 8 neurons, and an output layer of 2 neurons, comprising a total of 58 parameters. \rev{Due to our limited dataset size, we chose a simple model to reduce the risk of overfitting. Initial tests showed that adding layers or neurons did not improve accuracy and instead led to overfitting, capturing noise over useful patterns. This simpler structure provided computational efficiency while effectively capturing essential data patterns.} Both the input and hidden layers use the LeakyReLU activation function to introduce non-linearity, while the output layer uses a linear activation function to produce the final predictions.
For training, we use the Adam optimizer with a learning rate of \num{5e-4}\rev{, determined through systematic testing of values ranging from \num{e-2} to \num{e-6}. This learning rate provided stable convergence behavior and minimized training time without sacrificing model performance.} The loss function used is \gls{mse}. We use a batch size equal to the entire dataset size, performing batch gradient descent for each training step. This approach reduces the impact of noisy data points and is feasible given the limited size of our dataset.

\subsection{Model Validation and Performance Assessment Metrics}
\subsubsection{One-Step Prediction Errors}
At each time step in the test set $\mathbf{\mathcal{D}}^{test}$, generated from a different run \revdel{in}\rev{on} the same track, the identified models generate predictions for the subsequent states, $\mathbf{\hat{x}}_{k+1} = f(\mathbf{x}_k, \mathbf{u}_k)$. These predictions are then compared with the observed subsequent states, $\mathbf{x}_{k+1}$, using the \gls{rmse} to evaluate prediction accuracy.

\subsubsection{Racing Performance}
\revdel{Additionally,}We will \rev{also }evaluate the lap time to complete a lap\revdel{, and} \rev{as well as} the average and maximum lateral \revdel{lateral} deviation from the racing trajectory achieved by a model-based lateral controller \cite{map} utilizing the identified model parameters. These metrics serve as a practical performance measure in a real-world scenario, indicating how well the model supports the controller in achieving high racing performance. 

\subsection{Simulation Results}
\subsubsection{Experimental Setup}

The simulation experiments are conducted in the F1TENTH simulator \cite{forzaeth}, which \rev{was adapted to use the single-track vehicle model presented in \Cref{subsec:vehicle_model}} \revdel{uses the dynamic single-track vehicle model} with parameters similar to those of the real vehicle to minimize the sim-to-real gap. The simulator also provides ground truth data.

To benchmark our approach, we also employ the \gls{nls} method for comparison\rev{, applying the same low-pass filtering to the data for both approaches to ensure a fair comparison}. The objective function for \gls{nls} aims to find the best Pacejka parameters \(\mathbf{\Phi_p}\) that minimize the sum of squared residuals between the observed and predicted states over all time steps and is formulated as follows:
\begin{equation} \label{eq:nls_obj_func}
\begin{aligned}
\min_{\mathbf{\Phi_p}} \sum_{k=1}^{N} & \alpha_{v_y}^2 \left( v_{y,k+1} - \hat{v}_{y,k+1} \right)^2 + \alpha_{\omega}^2 \left( \omega_{k+1} - \hat{\omega}_{k+1} \right)^2.
\end{aligned}
\end{equation}
\rev{The scaling parameters $\alpha_{v_y},\,\alpha_{\omega}$ were both set to 1 throughout this work.} For both approaches, 30 seconds of driving data, corresponding to 2 laps on the track, with a time step of 0.02 second\rev{s} is collected within the simulator by driving the vehicle with a model-free Pure Pursuit \cite{pure_pursuit} controller to follow the predefined trajectory. 

Given the simulator's simplified vehicle model and the availability of ground truth data, both methods are expected to yield accurate models. To evaluate the robustness of each method under noisy conditions, we introduce artificial Gaussian noise to the data, simulating real-world measurement noise. The procedure for adding noise is as follows:

\begin{equation}
\mathbf{d}_{\text{noisy}} = \mathbf{d} + \mathcal{N}(0, \mathbf{d}_{\text{avg}} \cdot \eta),
\end{equation}
where \( \mathbf{d} = [v_x, v_y, \omega, \delta] \) represents the state and input variables, \( \mathbf{d}_{\text{avg}} \) are their respective average magnitudes to maintain a consistent noise-to-signal ratio across different levels of the state and input variables. \(\mathcal{N}(0, \sigma)\) denotes Gaussian noise with zero mean and standard deviation \(\sigma\), scaled by the noise multiplier \(\eta\), which is increased gradually after each experiment from 0 to 1.4 with an increment of 0.2.

\subsubsection{Model Validation Testing}

\begin{figure}[h!]
    \centering
    \includegraphics[width=\columnwidth]{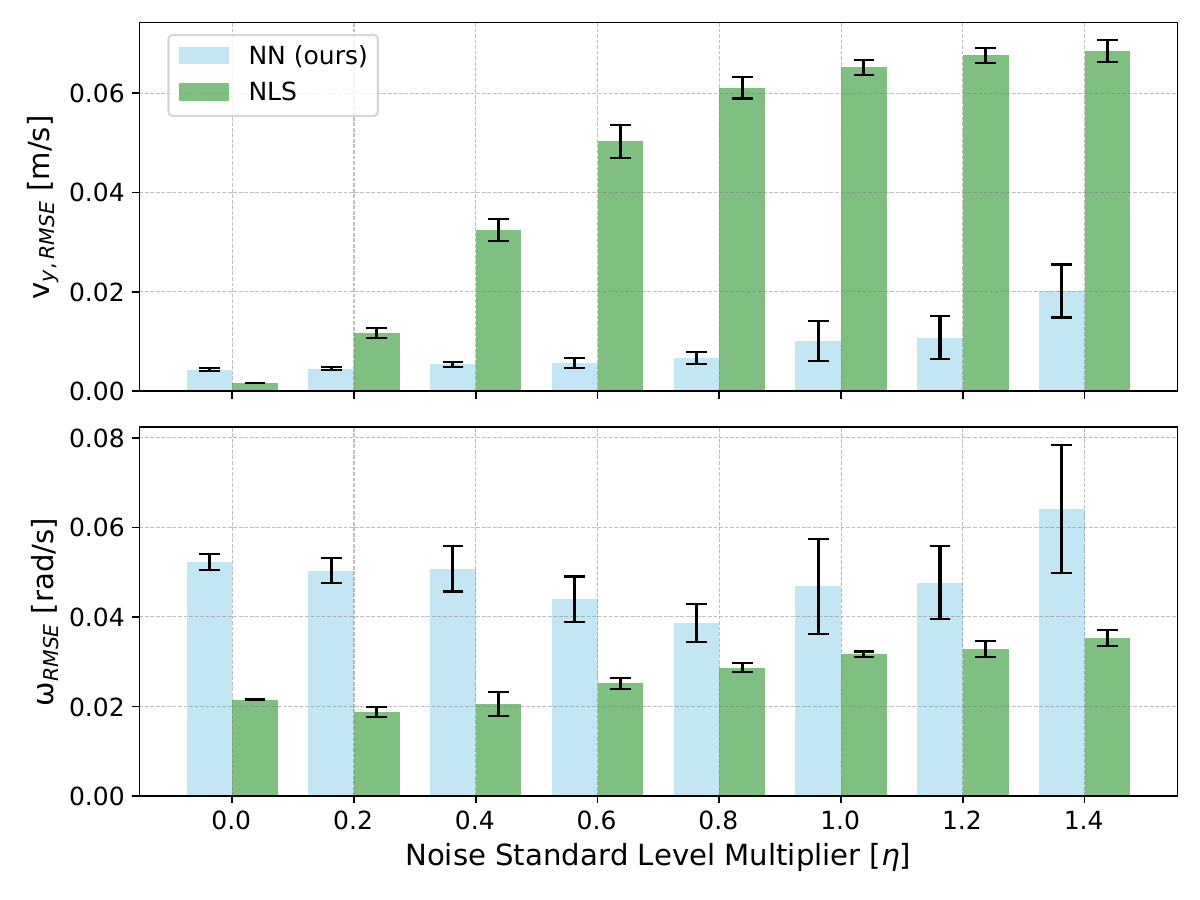}
    \caption{The \gls{rmse} of lateral velocity \rev{and yaw rate }predictions for \rev{the }models identified \revdel{with}\rev{using} our approach and the \gls{nls}\revdel{ method} on the test set under \revdel{varying}\rev{increasing} levels of noise. \rev{Experiments were repeated 10 times.}}
    \label{fig:ours_nls_noise}
\end{figure}

\Cref{fig:ours_nls_noise} illustrates the \gls{rmse} of lateral velocity \rev{($v_y$) and yaw rate ($\omega$)} predictions under varying levels of noise. The x-axis represents the noise standard deviation multiplier ($\eta$), which scales the standard deviation of \rev{the }Gaussian noise added to the data. \revdel{The y-axis represents the \gls{rmse} in meters per second (m/s) for each identified model. }

\rev{As our initial tests showed that the identified Pacejka models typically converge within six iterations across all noise levels, we run our approach for six iterations.}\revdel{In the noiseless (\(\eta\) = 0) and low noise (\(\eta\) = 0.2) conditions, both approaches achieve high accuracy, demonstrating their effectiveness under ideal conditions with ground truth data. Across all noise levels, our method consistently exhibits lower \gls{rmse} values compared to the \gls{nls} method. As the noise standard deviation multiplier (\(\eta\)) increases, the \gls{rmse} of the \gls{nls} method rises sharply, whereas the proposed method shows a much slower increase.} \rev{For lateral velocity ($v_y$), our approach consistently outperforms the \gls{nls} method across all noise levels ($\eta$). In terms of yaw rate ($\omega$), the performance of our approach is comparable to that of the \gls{nls} method. When averaging the performance across both $v_y$ and $\omega$, the proposed method achieves an \gls{rmse} that is 3.3 times lower for the one-step prediction compared to the \gls{nls} method}\revdel{ Overall, experimental results show that our approach achieves an \gls{rmse} that is \rev{3.3} times lower than that of the \gls{nls} method on average}, highlighting its robustness and accuracy in handling noisy conditions.

\subsection{Physical F1TENTH Results}

\subsubsection{Experimental Setup}

For the real-world experiments, we used an F1TENTH vehicle \revdel{based on, with the setup }\rev{as} detailed in \cite{forzaeth}. This setup includes an onboard Intel NUC computer running all software algorithms within the ROS framework, enabling indoor localization and state estimation, as in \cite{forzaeth}.

In practice, due to the absence of a ground truth model, we use the widely adopted steady-state system identification \cite{ss_sys_id}, as a baseline. This method is expected to yield a highly accurate model because the Pacejka parameters are identified under controlled, simplified steady-state conditions. Additionally, the \gls{nls} method, with the objective function \eqref{eq:nls_obj_func}, is used as another on-track system identification benchmark to compare with the proposed approach.

\subsubsection{Data Collection}

\revdel{For steady-state system identification, we conducted three experiments with the vehicle driven at constant velocities and ramp steering angles over 30 seconds per experiment. Steering angles ranged from 0.2 to 0.4 radians, chosen based on the F1TENTH vehicle's steering limits and space constraints, as lower angles result in larger trajectories. The selected velocities were 2, 2.5, and 3 m/s, balancing the need for sufficient slip angle data and the available laboratory space. These maneuvers covered an area of 4.5m x 4.5m, but to ensure safe driving and avoid potential crashes, the total required free space was approximately 6.5m x 6.5m. These three experiments sufficiently covered a comprehensive range of the slip angles versus force curve.}
\rev{For steady-state identification, three experiments were conducted at constant velocities (2-\SI{3}{\metre\per\second}) and ramp steering angles (0.2–0.4 radians), covering a range of slip angles in a controlled 6.5m x 6.5m space.}

\rev{For both the NLS and our approach}\revdel{For on-track system identification}, 30 seconds of driving data\rev{---}equivalent to approximately 3 laps\rev{---}was collected on the same surface used for steady-state experiments. During these trials, the vehicle was driven to follow a predefined trajectory using a model-free Pure Pursuit \cite{pure_pursuit} controller, assuming no prior knowledge of tire parameters. This method provided diverse and dynamic data, crucial for on-track system identification. The choice of 30 seconds was based on preliminary tests\rev{, which} show\rev{ed}\revdel{ing} it \revdel{provided enough data}\rev{was sufficient} for robust parameter identification.

\subsubsection{Model Validation Testing}

\begin{figure}[h!]
    \centering
    \includegraphics[width=\columnwidth]{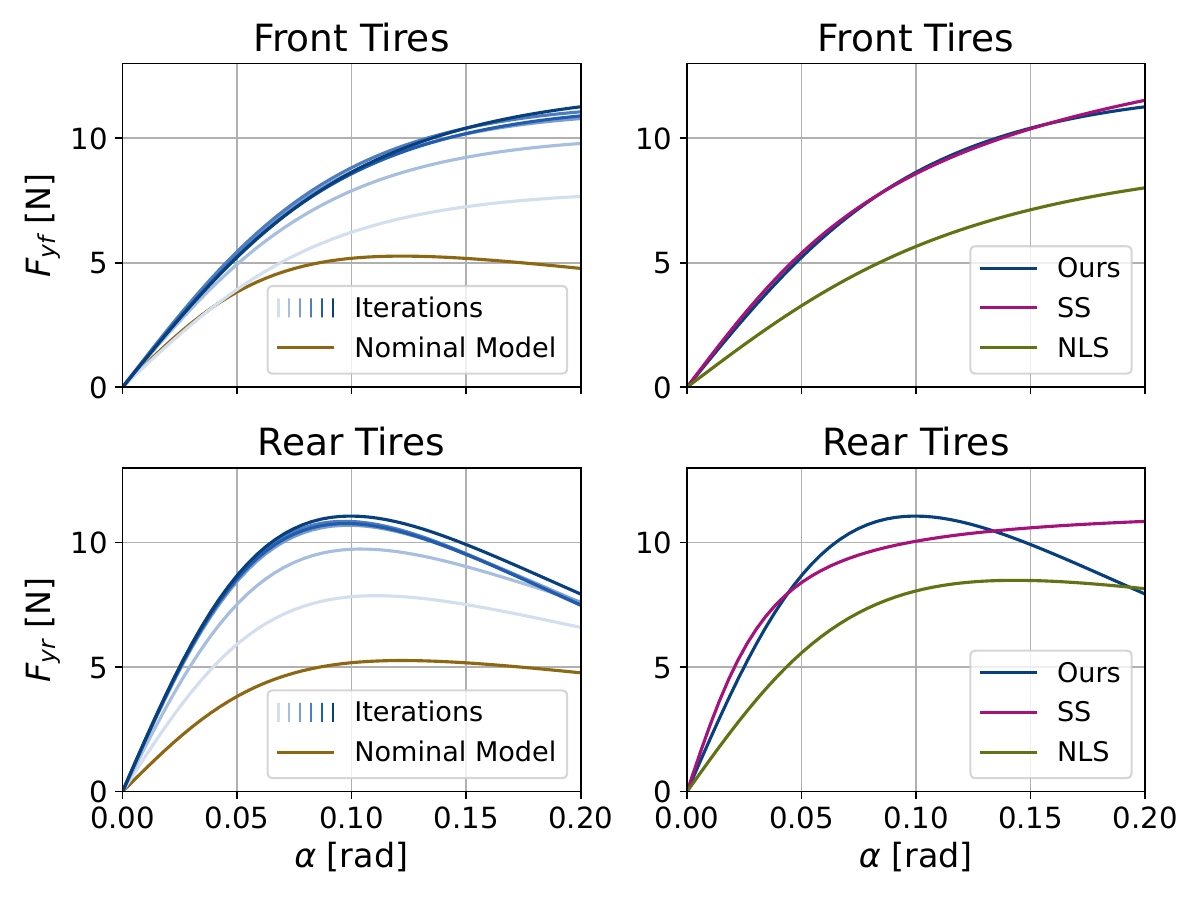}
    \caption{\rev{The left plots illustrate the evolution of the resultant Pacejka models using our method over six iterations, starting from the nominal model (bronze) and transitioning through shades of blue (light to dark). The right plots show the identified Pacejka models using three methods: Ours, Steady-State (SS), and \gls{nls}.}}
    \label{fig:identified_models_real}
\end{figure}

\revdel{Using on-track data, we conducted both the \gls{nls} approach and our method, while steady-state data was used for steady-state system identification. }
\Cref{fig:identified_models_real} shows the identified Pacejka models for both front and rear tires using the three approaches. The models identified with our approach align well with the baseline steady-state method, while the \gls{nls} \revdel{approach }fails to produce similar models. At higher lateral slip angles\revdel{, }\rev{---}greater than 0.15 and 0.13 radians for the front\revdel{ tire} and rear tire\rev{s}, respectively\revdel{, }\rev{---}the models diverge due to insufficient data in that region\rev{,} as it \revdel{is above}\rev{lies outside} the operational region. Therefore, our region of interest is limited to slip angles below 0.15 and 0.13 radians for the front and rear tires, respectively. Within this region, both our approach and the steady-state method show perfect alignment for the front tires, while slight deviations for the rear tires are observed, likely due to noise in the data.

\rev{\Cref{tab:real_results} presents a comparison of the three methods using key metrics, including the \gls{rmse} for lateral velocity ($v_y$) and yaw rate ($\omega$) on the test data, as well as lap time, average lateral error, and maximum lateral error. All runs were performed with a \gls{map} controller \cite{map,forzaeth}, utilizing the tire parameters identified by each method. Despite being considered the gold standard, the steady-state method shows the worst \gls{rmse} performance for $\omega$ on a noisy real-world dataset, as shown in \Cref{tab:real_results}. Similarly, although the \gls{nls} achieves the lowest \gls{rmse} for $\omega$, the resulting Pacejka parameters underestimate tire forces, as shown in \Cref{fig:identified_models_real}, leading to undrivable (i.e., N.C.) behavior in closed-loop tests.}

\revdel{\Cref{tab:real_results} presents a comparison of the performance and \gls{rmse} values for the three methods with the key metrics including the \gls{rmse} for lateral velocity $v_y$ and yaw rate $\omega$ on the test data, as well as lap time, average lateral error, and maximum lateral error. All runs were performed with a model-based \gls{map} controller \cite{map,forzaeth}, utilizing the identified tire parameters of each identification method. Our approach demonstrates superior performance with lower \gls{rmse} for $v_y$ and faster lap times, indicating better overall accuracy and efficiency. The steady-state method shows slightly higher \gls{rmse} values and lap times, while the \gls{nls} approach, despite having the lowest \gls{rmse} for $\omega$, fails to complete a lap without track violations, resulting in incomplete (N.C.) data for lap time, average lateral error, and maximum lateral error.}
\begin{table}[h]
\vspace{0.3cm}
\centering
\begin{tabular}{c|c|c|c}
\toprule
\textbf{Metrics} & \textbf{Ours} & \textbf{Steady-State} & \textbf{NLS} \\ \midrule
$v_{y,RMSE}$ [m/s]  & \textbf{0.0233} & 0.0239 & 0.0289 \\
$\omega_{RMSE}$ [rad/s]  & 0.106 & 0.108 & \textbf{0.0655} \\ \midrule
$t^\mu_{lap}$ [s] & \textbf{8.10} & 8.16 & N.C. \\
$\vert d^\mu \vert$ [m] & \textbf{0.082} & 0.092 & N.C. \\
$\vert d^{max} \vert$ [m] & \textbf{0.27} & 0.3 & N.C. \\ \bottomrule
\end{tabular}
\caption{The \gls{rmse} values for lateral velocity $v_y$ and yaw rate $\omega$, along with lap times, average, and maximum deviation from the trajectory for different methods: our approach, the steady-state system identification method, and the \gls{nls} approach. N.C.: Not completed a lap without track violations}
\label{tab:real_results}
\end{table}

The experimental results demonstrate that our proposed on-track system identification method accurately identifies tire parameters, producing Pacejka models similar to those obtained from traditional steady-state methods, particularly within the critical slip angle range. Our approach offers significant operational simplicity compared to steady-state system identification and greater robustness compared to the classical \gls{nls} method while achieving the best \revdel{results}\rev{racing performance} among the benchmark methods. 

Most notably, our approach require\revdel{d}\rev{s} only six iterations, with each iteration taking approximately 0.5 seconds, totaling \rev{just} 3 seconds to learn the complete \rev{vehicle} model\rev{, even }without \rev{accurate }prior knowledge.\revdel{ This rapid convergence\revdel{ not only} highlights the efficiency of our method\revdel{ but also its potential for real-time adaptation}, enabling dynamic adjustments during racing scenarios every 0.5 seconds. \revdel{These findings }} \rev{This} confirm\rev{s} the accuracy, efficiency, and practicality of our method for real-world autonomous racing applications. \revdel{, making it a powerful tool for enhancing performance in highly dynamic environments.}

\rev{\subsection{Dynamic Adaptation}}
\rev{This experiment evaluates our method’s ability to adapt from an initial hard tire model to a soft tire model under real-world conditions, such as tire softening due to temperature changes. A controller based on the hard tire model is employed while the vehicle operates with soft tires. Since only a slight model mismatch is assumed, our method is executed only for two iterations, completing in a total of 1 second. As shown in \Cref{tab:hard_soft_tire_exp}, adapting to the soft tire model achieves a lap time of 8.5 seconds and an average lateral error of 0.11 meters, outperforming the initial hard tire model by 0.25 seconds and reducing the lateral error by 0.05 meters.}

\begin{table} [h]
\centering
\begin{tabular}{c|c|c}
\toprule
\textbf{Metrics} & \textbf{Hard Tire Model} & \textbf{Adapted Soft Tire Model} \\ \midrule
$t^\mu_{lap}$ [s] & 8.75 & \textbf{8.5} \\ 
$\vert d^\mu \vert$ [m] & 0.16 & \textbf{0.11} \\ \bottomrule
\end{tabular}
\caption{\rev{Performance comparison showing adaptation from an initial hard tire model to a soft tire model, reflecting real-time responsiveness to changes in tire conditions.}}
\label{tab:hard_soft_tire_exp}
\end{table}

\rev{These results demonstrate our method's ability to adapt in real-time within just 1 second, utilizing 30 seconds of driving data, making it suitable for dynamic racing conditions that require rapid adjustments.}

\section{CONCLUSIONS}
\revdel{In this paper, we propose a novel learning-based on-track system identification method for high-performance autonomous racing vehicles, integrating a \gls{nn} with a nominal Pacejka tire model. The proposed approach offers several key advantages.
Firstly, it eliminates the need for large, open spaces required by traditional steady-state methods by utilizing on-track data, simplifying the identification process and making it more practical for real-world applications. Secondly, it achieves up to \rev{3.3} times lower \gls{rmse} with increased noise levels compared to the \gls{nls} method, ensuring accurate identification even in noisy conditions.
Thirdly, it uses three times less data than the steady-state method, requiring only 30 seconds of training data and 3 seconds of training time, totaling 33 seconds to fully learn the model without prior knowledge. Lastly, it adapts in a single iteration within 0.5 seconds for slight model changes, demonstrating its capability for real-time updates.
The identified Pacejka models closely match those from traditional steady-state methods within the critical slip angle range, highlighting the operational simplicity, robustness, and efficiency of the proposed approach. This makes the method a practical solution for autonomous racing, enhancing the adaptability and performance of vehicles under varying conditions and advancing the state-of-the-art in autonomous racing.}\rev{This paper presents a learning-based on-track system identification method for autonomous racing, integrating a \gls{nn} with a nominal Pacejka tire model. The approach eliminates the need for large, open spaces such as required by traditional steady-state methods by utilizing on-track data, simplifying the process, and achieving up to 3.3 times lower \gls{rmse} under noise compared to standard \gls{nls} methods. It uses three times less data than steady-state methods, learning the model in 33 seconds, and adapts in 1 second for environmental changes, enabling real-time updates. The identified Pacejka models closely align with steady-state methods within critical slip angles, demonstrating its practicality, robustness, and efficiency for advancing autonomous racing performance.}\revdel{Future work will focus on extending the method to identify more complex vehicle models and testing its applicability to larger scale vehicles.}
\rev{Different lines of future work are possible. One direction is to improve the accuracy of the nominal model by precisely refining the tire model and then combining it with more complex \gls{ml} models, such as those presented in \cite{xiao2024anycaranywherelearninguniversal, kalaria2024agilemobilityrapidonline}. Another direction involves exploring metrics beyond one-step prediction \gls{rmse} that could better assess the performance of the identified Pacejka models and help define a stopping criterion, as the current metric has proven to be not entirely indicative of Pacejka parameters identification performance. Furthermore, given its small computational footprint, the full \gls{nn} corrected vehicle model could be used in the context of a simulator, similarly to \cite{chrosniak2024deepdynamics}, or in a sampling-based \gls{mpc} method as in \cite{kalaria2024agilemobilityrapidonline}. Finally, further extensions could also consider applying the same algorithm to other robotic platforms, where system identification would require inconvenient infrastructure.}


\section*{ACKNOWLEDGMENT}
The authors would like to thank all team members of the \emph{ForzaETH} racing team. In particular Niklas Bastuck for the photography of the live car in \Cref{fig:graphical_abstract}.


\bibliographystyle{IEEEtran}
\bibliography{main}

\end{document}